# New PCA-based Category Encoder for Cybersecurity and Processing Data in IoT Devices

Hamed Farkhari[*†], Joseanne Viana [†‡], Luis Miguel Campos [*], Pedro Sebastião [†‡],
Luis Bernardo [§‡]
[*]PDMFC, Rua Fradesso da Silveira, n. 4, Piso 1B, 1300-609, Lisboa, Portugal
[†]ISCTE – Instituto Universitário de Lisboa, Av. das Forças Armadas, 1649-026 Lisbon, Portugal
[‡]IT – Instituto de Telecomunicações, Av. Rovisco Pais, 1, Torre Norte, Piso 10, 1049-001 Lisboa, Portugal
[§]FCT – Universidade Nova de Lisboa, Monte da Caparica, 2829-516 Caparica, Portugal;
Emails : Hamed_Farkhari@iscte-iul.pt, joseanne_cristina_viana@iscte-iul.pt, luis.campos@pdmfc.com,
pedro.sebastiao@iscte-iul.pt, lflb@fct.unl.pt

*Abstract*—Increasing the cardinality of categorical variables might decrease the overall performance of machine learning (ML) algorithms. This paper presents a novel computational preprocessing method to convert categorical to numerical variables ML algorithms. It uses a supervised binary classifier to extract additional context-related features from the categorical values. Up to two numerical variables per categorical variable are created, depending on the compression achieved by the Principal Component Analysis (PCA). The method requires two hyperparameters: a threshold related to the distribution of categories in the variables and the PCA representativeness. This paper applies the proposed approach to the well-known cybersecurity NSLKDD dataset to select and convert three categorical features to numerical features. After choosing the threshold parameter, we use conditional probabilities to convert the three categorical variables into six new numerical variables. After that, we feed these numerical variables to the PCA algorithm and select the whole or partial numbers of the Principal Components (PCs). Finally, by applying binary classification with ten different classifiers, we measure the performance of the new encoder and compare it with the other 17 well-known category encoders. The new technique achieves the highest performance related to *accuracy* and *Area Under the Curve* (*AUC*) on high cardinality categorical variables. Also, we define the harmonic average metrics to find the best trade-off between train and test performances and prevent underfitting and overfitting. Ultimately, the number of newly created numerical variables is minimal. This data reduction improves computational processing time in Internet of things (IoT) devices in future telecommunication networks..

## I. INTRODUCTION

Machine learning (ML) prediction problems require giving the model relevant features to represent the problem accurately. Consequently, data preparation and feature engineering are critical activities for all machine learning algorithms [1]. In Internet of Things (IoT) devices: processing capacity, energy consumption, and resource availability all limit the execution of deep learning algorithms. One way to overcome these restrictions is to use high-performance algorithms that consume fewer resources. As the amount of accessible data rises, the degree of diversity of the features increases and this expansion impacts categorical variables. When the number of features grow, the cardinality, which is the number of unique values detected in each feature, increases [2]. The challenge of appropriately and effectively encoding categorical features influence the machine learning model's performance. Handling the conversion from categorical characteristics to numerical features is a well-known issue in data science and machine learning since many methods require numerical input [3]. This problem has several solutions. Specific categorical data encoding schemes are more suitable than others depending on the type of problems i.e. classification or regression. These encoders are critical when processing large volumes of data, especially in IoT devices, at the edge, and in cloud computing because errors and outliers are more common when using these devices to process data. Due to these errors and outliers, reliable statistical estimations are challenging to compute.

One Hot Encoding is the most well-known encoding for low-cardinality categorical features. This yields orthogonal and equidistant vectors for each of the categories. Integers are picked at random because they have no inherent order. An alternate encoding method is Label/Ordinal Encoding, which uses a single column of integers to represent multiple category values. Both encoding techniques present high-dimensional encoding limitations, but Label Ordinal Encoding forces the categories into a particular order. This makes it more difficult for the model to extract valuable information. Regarding the assessment of ML algorithms' success, researchers have used a variety of methodologies such as Recall, Precision, F-Factor area under the curve, true positive rate (TPR), true negative rate (TNR) and *accuracy* [4]. In most cases, the focus is on specific attributes that matter in the context for which the measure was developed. For example, when Information Retrieval (IR) algorithms are evaluated on Recall, Precision, and F-Factor, erroneous predictions are often overlooked in favor of the accurate ones.

We present a different approach to solving the categorical encoders' modelling problem using conditional probability in supervised learning and Principal Component Analysis (PCA). Further, we compare the performance of our method with several available categorical encoders and classifiers using the same dataset. Finally, we show that our method achieves the best performance by adjusting only two parameters. Our algorithm, which outperforms current machine learning algorithms and reduces the dimensionality of the data, may be a viable choice for IoT devices and cybersecurity algorithms embedded in sensors, UAVs, and other devices. This paper is organized as follows: First, we introduce our novel method to convert categorical to numerical variables considering the probability relationships between categories and target classes in supervised classification. Then, we add a description of the metrics that select the best combinations between them and propose a new metric based on the harmonic averages to highlight the improvements in *accuracy* during training. After that, we analyze the results from 17 different categorical to numerical encoders using ten different classifiers. We compare the results using the *accuracy*, the *Area Under the Curve* (*AUC*) and the *harmonic averages*, highlighting the improvements in *accuracy* during training.

*A. Contributions and Motivation*

Traditional categorical encoders do not provide the parameters to adjust them to the classifiers. Considering this constraint, below is a summary of the main contributions of this paper:

- A new method to encode categorical features using only two hyperparameters: a combination of threshold and PCA that adjusts to different classifiers for maximum performance achievement.
- A supervised category encoder which is suitable for both linear and nonlinear classification algorithms.
- A new metric for measuring training gains in *accuracy* using *Harmonic Averages* calculations.
- A comparison between the proposed solution and the available categorical encoders using *accuracy*, *AUC* and the proposed metric based on *harmonic averages*.

In high cardinality categorical variables, our method achieves the highest performance using the lowest possible dimensionality, specifically, when the categories exist in the test set and not in the train set. Furthermore, it is possible to prevent or decrease underfitting and overfitting. Also, we define new metrics using a different set of hyperparameters which makes adjustments in the classifiers during the preprocessing steps to improve the performance of our encoder.

## II. THE PROPOSED METHOD

The scheme offers a unique computational preprocessing approach for converting categorical to numerical variables for machine learning (ML) methods. Table I shows the dataset with categorical variables named from $Variable_1$ to $Variable_N$ and each variable contains different numbers of categories. It is required that the target variable defines a binary classification, with two complementary classes.

| $Variable_1$ | ... | $Variable_N$ | Target |
|---|---|---|---|
| $Category_{1,1}$ | ... | $Category_{N,1}$ | |
| $Category_{1,2}$ | ... | $Category_{N,2}$ | Class $C_1$ or $C_2$ (Binary Classification) |
| . | ... | . | |
| $Category_{1,j}$ | ... | $Category_{N,j}$ | |
| . | | . | |

TABLE I: Categorical variables with different categories in binary classification.

Using the variables and the target in Table I, we define the conditional probabilities for each unique category using binary classification. The calculation for each category is based on the numbers of its occurrences for each class $C_1$ and $C_2$ per its total occurrences as (1) and (2) illustrates:

$$P1_{i,j} = P(Target = C_1 | Variable_i = Category_{i,j}), \quad (1)$$

$$P2_{i,j} = P(Target = C_2 | Variable_i = Category_{i,j}). \quad (2)$$

Before applying the threshold parameter, for each unique $Category_{i,j}$ the following condition holds:

$$P1_{i,j} + P2_{i,j} = 1, \quad (3)$$

where $i, j$ are defined as $\forall i, j | i \in \{1, 2, \ldots, N\}, j \in \{1, 2, \ldots, M_i\}$, $N$ is the number of total categorical variables, and $M_i$ is the number of unique categories for variable $i$. $N$ and $M_i$ are fixed for each variable. Thus, each category $Variable_i$ will produce two new numerical variables with three states.

| $Variable_i$ | New $Var1_i$ | New $Var2_i$ | Conditions |
|---|---|---|---|
| $Category_{i,j}$ | 1 | 0 | If $P1_{i,j} > P2_{i,j}$, AND $P2_{i,j} > threshold$. |
| $Category_{i,j}$ | 0 | 1 | If $P1_{i,j} < P2_{i,j}$, AND $P1_{i,j} > threshold$. |
| $Category_{i,j}$ | 0 | 0 | If $P1_{i,j} < threshold$, OR $P2_{i,j} < threshold$. |

TABLE II: Converting each categorical variable to two numerical variables with conditions for each category.

In Table II, $Variable_i$, $NewVar1_i$ and $NewVar2_i$ refers to categorical $Variable_i$, and the first and second newly created numerical variables for $Variable_i$, respectively. New numerical variables will be created based on the probability conditions in (1), (2), and the threshold value. Each categorical value of a database element is converted to the $NewVar1_i$ and $NewVar2_i$ values, where the elements' Category is used to select the value of $j$ in Table II.

*A. Threshold*

The *threshold* defines the first hyperparameter, which specifies a minimum occurrence probability for a category considered in the binary classification. Probabilities $P1_{i,j}$ and $P2_{i,j}$ are calculated using (1) and (2) based on the classification of $C_1$ and $C_2$ of the database samples. Our method creates two new numerical variables for each categorical variable using the equations specified in Table II. Categories with rare elements from one class (with a probability below the threshold) are mapped into ($NewVar1$, $NewVar2$) = (0, 0). Otherwise, the variables contain the majority class, $C_1$, ($NewVar1$, $NewVar2$) = (1, 0), or $C_2$, ($NewVar1$, $NewVar2$) = (0, 1).

*B. Principal Component Analysis*

The second hyperparameter is the number of Principal Components (PCs) available after the PCA processing. The main objective of PCA in our methodology is to remove the correlation between the $2N$ new numerical features in Table II, where $N$ defines the number of categorical variables. The number of PCs, denoted as $K$, can vary from 1 to $2N$. $K$ can be the minimum number of PCs necessary to capture all data variances, which might be below $2N$ if some numerical variables contain only one unique value for all categories (i.e., only ones or zeroes) or can be written as combinations of other numerical variables. By choosing a lower $K$, the cumulative data variance will be less than one. We describe the variety of the first and second hyperparameters in the grid search section.

*C. Scaling*

Usually, scaling is applied before PCA to prevent the feature dominance effect where some features overshadow others because they have different scales. In our method, there is no need for scaling because the new numerical features are normalized between zero and one. However, after the PCA process, the standardization scale using mean and standard deviation is applied for faster convergence in some classifiers, such as Support Vector Machines (SVMs).

*D. Dataset*

We choose the NSLKDD dataset [5] to test different encoding methods and classifiers because it is common in cybersecurity research (for instance, for network intrusion detection). The NSLKDD is divided into four different partitions: KDDTrain+, KDDTrain+_20Percent, KDDTest+, and KDDTest-21. All partitions are available for downloading from [5]. We use the KDDTrain+ exclusively for training, and the KDDTest+ as a complete test dataset for test purposes which includes all the test instances. A quick analysis of the NSLKDD shows that the KDDTrain+, KDDTest+, KDDTrain+_20Percent, and KDDTest-21 contain 125973, 22544, 25192, and 11850 samples, respectively. There are only three categorical variables in the dataset namely: protocol_type, service, and flag. We convert the categorical variables to numerical using different encoders to compare the performance of each method in binary classification.

*E. Categorical Encoders Dimensionality*

One the the main challenges related to high cardinality categorical variables is their high dimensionality after converting them to numerical features. The One Hot Encoding method presents such constraints. In our proposed method, the number of dimensions of new numerical features varies from a range of one to six. The protocol_type and flag variables in both KDDTrain+ and KDDTest+ sets contain the same cardinality. However, the cardinality of the service variable is greater and different between the train and test sets which may lead to a low performance of the available encoder. Table III shows the differences of dimensionality for newly created numerical features for each of the encoding schemes. The categorical encoders used are from the category_encoders library version 2.2.2. According to Table III, other encoder schemes create at least three dimensions for the new numerical features. In our system, it is possible to reduce them to one.

*F. Classifiers*

We use ten classifiers with different configurations in Python v3.6.9 and Sci-kit learn library v0.23.2 to compare the results. Table IV presents the classifiers with hyperparameters. For replication purposes, the seed value of randomness (*random_state*) in all classifiers is zero.

*G. Metrics*

Metrics such as *accuracy* can simply be measured in multi-class problems. However, other metrics such as precision, recall, FPR, F1-Score, and the sum of the Area Under the Curve

| Encoding Scheme (abbreviation) | Dim. |
|---|---|
| **(Proposed)** | **1-5** |
| Backward Difference Encoder (Backward Difference) [6] | 81 |
| BaseN Encoder (BaseN) [7] | 13 |
| Binary Encoder (Binary) [8] | 13 |
| Cat Boost Encoder (Cat Boost) [9] | 3 |
| Count Encoder (Count) [10] | 3 |
| Generalized Linear Mixed Model Encoder (GLMM) [11] | 3 |
| Hashing Encoder (Hash) [12] | 8 |
| Helmert Encoder (Helmert) [6] | 81 |
| James-Stein Encoder (James-Stein) [13] | 3 |
| Leave One Out Encoder (LOOE) [14] | 3 |
| M-estimate Encoder (MEestimate) [15] | 3 |
| One Hot Encoder (One Hot) [6] | 84 |
| Ordinal Encoder (Ordinal) [6] | 3 |
| Polynomial Encoder (Polynomial) [6] | 81 |
| Sum Encoder (Sum) [6] | 81 |
| Target Encoder (Target) [15] | 3 |
| Weight of Evidence Encoder (WOE) [16] | 3 |

TABLE III: Comparing dimensionality of new numerical features created by each Encoding scheme.

| Classifiers | hyperparameters |
|---|---|
| Logistic Regression (LR) | solver = 'saga', penalty = 'l2', c = 1.0 |
| Multilayer Perceptron (MLP) | solver = 'adam', alpha = 0.0001, hidden_layer_sizes = 100, activation = relu, learning_rate_init = 0.001('constant'), batch_size=200 |
| SVM 1 | kernel = rbf, gamma = 'auto', c=1.0 |
| SVM 2 | kernel = poly, gamma='auto', c=1.0, degree=5 |
| SVM 3 | kernel = linear, c=1.0 |
| Decision Tree(DT) | max depth=5, split quality measure = 'gini', max features considered for each best split = min(8, number of new numerical features) |
| Ada Boost Classifier (ADA 1) | base_estimator=DecisionTreeClassifier (max_depth=1), n_estimators=50 |
| Ada Boost Classifier (ADA 2) | base_estimator=DecisionTreeClassifier (max_depth=5), n_estimators=10 |
| Random Forest (Forest) | max depth =5, no. of estimators = 10, split quality measure = 'gini', max features considered for each best split = min (5, number of new numerical features) |
| Gaussian Naive Bays (GNB) | default sci-kit learn parameters |

TABLE IV: 10 Classifiers with hyperparameters used for classification.

(*AUC*) of the Receiver Operating Characteristic (ROC) cannot be easily calculated [17]. Thus, in practice, *accuracy* may be enough to check performance in multi-class problems. It is essential to choose the proper metrics to compare the results between the available encoders and the proposed system. We use binary classification and divide the target labels associated with attacks and regular Internet traffic (normal labels). The proportions of attack and normal labels in the train set is 46.54% and 53.46%. In the test set, the ratios are 56.92% and 43.08%, respectively. The percentages of labels in two of the classes show that the number of instances in the train and test sets are balanced. On one hand, balanced classification usually uses *accuracy* and *AUC*. On the other hand, unbalanced classification uses precision, F1-score, and other metrics.

## H. New Metrics

Commonly, the attacks and normal data in the train and test datasets are not equal. For example, the NSLKDD test set contains only 15% of the total data. The unbalanced test data has an impact on the evaluation of the algorithm's learning capabilities. Even if the test set exhibits an excellent performance, it does not guarantee that the same performance will occur in the training data set and vice versa. We should therefore consider a trade-off between the performances of train and test sets. The effect of changing the amount of data available for the test by 1% is less noticeable that in the train. If the *accuracy* of the algorithm changes 1% in the test, it affects only 15% of total data, for our data set is 22544 samples. Nevertheless, a 1% change in the training data affects the other 85% of data containing 125973 samples. For the first time, we want to define new metrics to consider both train and test performances because extensive changes may occur in the train when we ignore minimal changes in test performance. In cybersecurity, these changes mean our systems can detect more attacks, and protection increases. We define new metrics and compare our system's performance using both the previous and the new metrics in light of the above explanation. The new metrics are the distance to the ideal point as the error to calculate mean squares errors (MSE) and the harmonic average of the same metrics in the train and test sets. Using only one of these three metrics is adequate for sorting encoder performance and fine-tuning hyperparameters in our proposed encoder. In addition, using these metrics avoids overfitting or underfitting problems, which the following sections discuss. Equations 4, 5, and 6 use the new metrics to estimate performance:

- Mean Square Errors (MSE) to the ideal point for *accuracy*:

$$MSE = 0.5[(100-a)^2 + (100-b)^2]; \quad (4)$$

- Mean Square Errors (MSE) to the ideal point for *AUC*:

$$MSE = 0.5[(1-c)^2 + (1-d)^2]; \quad (5)$$

- Harmonic average of the same metrics (*accuracy* or *AUC*) in train and test:

$$Harmonic\_avg = \frac{(2.e.f)}{(e+f)}; \quad (6)$$

where in (4), $a$ and $b$ are percentage accuracies in train and test data. In (5), $c$ and $d$ are $AUCs$, for the same data. The harmonic averages in (6) defines $e$ and $f$ using *accuracy* or $AUCs$ in the data, respectively. The harmonic average is defined to calculate the average between train and test sets for the same metrics. We apply our method to the NSLKDD dataset containing three categorical variables. We use one unique threshold for all of them due to the similar distribution of the classes in the category of the three categorical variables. All threshold values are represented as percentages.

## III. EXPERIMENTAL RESULTS

We apply our method to the NSLKDD dataset containing three categorical variables. We use one unique threshold for all of them due to the similar distribution of the classes in the category of the three categorical variables. All threshold values are represented as percentages.

## A. Categorical Encoders Comparison

We measure the performance of a combination of 17 different encoders, plus ours from Table III, with the ten classifiers from Table IV, to compare our new proposed encoder algorithm with the other existing encoders. Table V identifies the 18 Encoders by their abbreviations and summarizes their performance results. Each column in the table V associates the encoding scheme with the best suitable classifier according to the train or test for *accuracy* or *AUC*. In the fourth column, we use the maximum *harmonic averages* of train and test *accuracies* to compare the results and sort the encoders from best to worst performance. For example, the test *accuracy* for the Polynomial encoder is 88.9549%, which is the highest *accuracy* that this encoder achieves using the GNB classifier. All the encoders are tested with all classifiers and table V presents the classifier with the highest performance. In our method, the hyperparameters Thre(1.87) and PCs(3) represents a threshold of 1.87 % and the top three principal components, respectively. Our algorithm achieves the highest test *accuracy* of 89.638041 % by feeding only the first principal component to the SVM2 classifier from Table IV, and with the two different thresholds of 3.64 % and 5.45 %. This *accuracy* is the highest out of all combinations of categorical encoders and classifiers and puts our encoder in the first place. Our method is placed second after Polynomial Contrast coding by choosing the Harmonic average of accuracies as a sorting metric, as is shown in Table V.

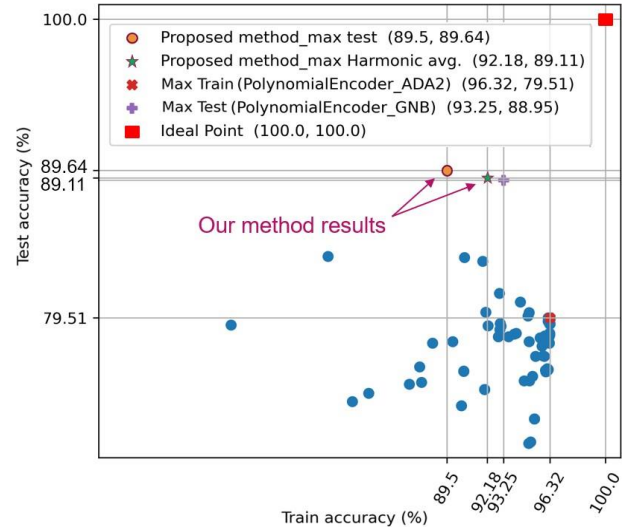

Fig. 1: Scatter of train vs test sets *accuracies* achieved by combination of 18 category encoders.

Figs 1 and 2 compare the *accuracies* and $AUCs$ of the 18 encoders with ten different classifiers with respect to the train versus test data. In fig 1, the point at [100, 100] represents the maximum train and test *accuracy* which is the ideal point of all encoders. After analyzing the available encoders, we discover that the polynomial achieves the greatest train and test *accuracies* with the ADA2(96.31%) and GNB(88.95%) classifiers. Our method achieves the highest **test** *accuracy* (89.64%) with the SVM2 classifier in comparison with the polynomial (88.95%).

Using the harmonic average of accuracies, our method achieves 89.11% during the test phase, which is still the highest test *accuracy*, but lower than the previous test results of 89.64% *accuracy*. However, the amount of train *accuracy* increases from 89.5% to 92.18% and we lose 0.53 % in the test. The difference between the prior test *accuracy* and the harmonic average of accuracies is +2.68 % in the train set and only -0.53% in the test set. The loss between the *accuracy* and the harmonic average metrics for the test set is so minimal and there are significant benefits in the training set which implies that the harmonic average of accuracies is a better metric choice. Fig 2 describes the results based on the *AUC* metric for the same encoders and classifiers. The ideal point is [1.0, 1.0] for the *AUCs* train and test sets. We show that the polynomial and our encoder performances present nearly the same results using the new and previous metrics considering the approximation of two floating points for the test set. The performance for both of them is 0.89 in the test. The difference is in the train in which the polynomial reaches 0.93 while our method achieves 0.92.

| Encoding Scheme | Classifiers with Max. Train *accuracy* (%) | Classifiers with Max. Test *accuracy* (%) | Classifiers with Max. harmonic avg. of *accuracies* (%) | Classifiers with Max. Train *AUC* | Classifiers with Max. Test *AUC* |
|---|---|---|---|---|---|
| Polynomial | ADA2, 96.3167 | GNB, 88.9549 | GNB, 91.0538 | ADA2, 0.9629 | GNB, 0.888 |
| **Proposed** | **All except GNB, Thre(11.9), PCs(1-5), 95.380756** | **SVM2, Thre(3.64, 5.45), PCs(1), 89.638041** | **SVM3, Thre(1.87), PCs(3), 90.6161** | **All except GNB, Thre(11.9), PCs(1-5), 0.953976** | **SVM2, Thre(3.64, 5.45), PCs(1), 0.893252** |
| Ordinal | ADA2, 96.3151 | LR, 83.388 | LR, 87.42 | ADA2, 0.9629 | LR, 0.8514 |
| One Hot | SVM1, 96.3127 | DT, 83.7252 | ADA2, 87.1133 | SVM1, 0.9628 | DT, 0.8274 |
| Sum | MLP, 96.3143 | ADA2, 79.5245 | ADA2, 87.1133 | MLP, 0.9628 | Forest, 0.814 |
| Target | ADA2, 96.3167 | ADA2, 79.5067 | ADA2, 87.1081 | ADA2, 0.9629 | ADA2, 0.808 |
| Backward Difference | ADA2, 96.315 | Forest, 80.6112 | ADA2, 87.1075 | ADA2, 0.9629 | Forest, 0.8165 |
| Helmert | ADA2, 96.3127 | Forest, 81.2145 | ADA2, 87.1038 | ADA2, 0.9628 | Forest, 0.822 |
| Base-N | SVM2, 96.3127 | GNB, 83.6542 | ADA2, 87.1035 | SVM2, 0.9628 | GNB, 0.8534 |
| Binary | SVM2, 96.3127 | GNB, 83.6541 | ADA2, 87.1035 | SVM2, 0.9628 | GNB, 0.8534 |
| James-Stein | ADA2, 96.3167 | ADA2, 79.4979 | ADA2, 87.1028 | ADA2, 0.9629 | ADA2, 0.808 |
| Cat Boost | ADA2, 96.3159 | ADA2, 79.4979 | ADA2, 87.1025 | ADA2, 0.9629 | ADA2, 0.808 |
| GLMM | ADA2, 96.3167 | ADA2, 79.4934 | ADA2, 87.1001 | ADA2, 0.9629 | GNB, 0.8148 |
| LOOE | ADA2, 96.3167 | ADA2, 79.4934 | ADA2, 87.1001 | ADA2, 0.9629 | ADA2, 0.8079 |
| WOE | ADA2, 96.3151 | ADA2, 79.4934 | ADA2, 87.0995 | ADA2, 0.9629 | ADA2, 0.8079 |
| Count | ADA2, 96.3143 | ADA2, 79.4535 | ADA2, 87.0752 | ADA2, 0.9628 | ADA2, 0.8074 |
| MEstimate | ADA2, 96.3159 | ADA2, 79.112 | ADA2, 86.8703 | ADA2, 0.9629 | ADA2, 0.8046 |
| Hash | MLP, 91.9959 | GNB, 77.8921 | GNB, 83.4566 | MLP, 0.917 | GNB, 0.792 |

TABLE V: 18 different encoders with the best classifier for each one, compared and sorted based on max harmonic average of accuracies. The amount of thresholds for our proposed method are in percentage.

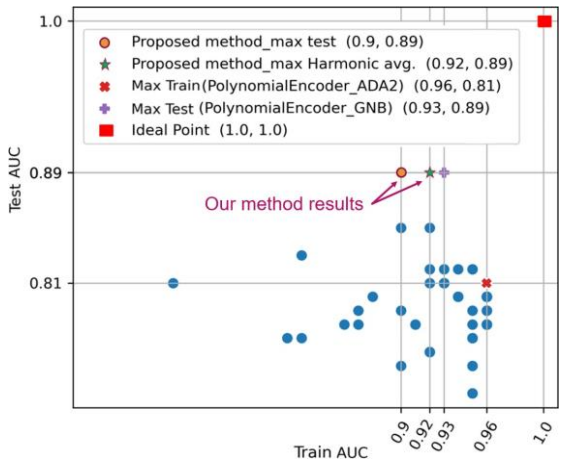

Fig. 2: Scatter of train vs test sets *AUC* achieved by combination of 18 category encoders.

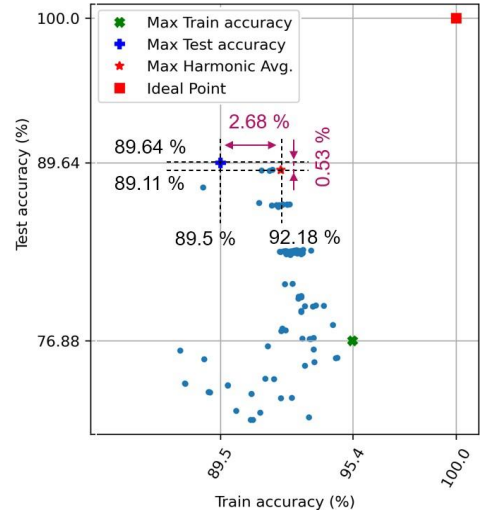

Fig. 3: Scatter of train vs test sets *accuracies*, grid search of two hyperparameters: Threshold and number of principal components followed by 10 different classifiers.

*B. Grid Search*

As previous sections describe, our new proposed category encoder contains two hyperparameters: the threshold and the number of principal components of the PCA. We conduct the grid search for all of the possible combinations of these two parameters to find the best values for each one. For the threshold, we check different values from 0.01% to 50%. Rare categories appear in either less than 1% or less than 5% of all instances. In our results rare categories occur a little more than five percent. We achieve the best test *accuracy* of 89.64% by choosing 5.45% or 3.64% as the threshold. We check all numbers in the threshold range together with different PC numbers that varies from 1 to 6 as the second hyperparameter.

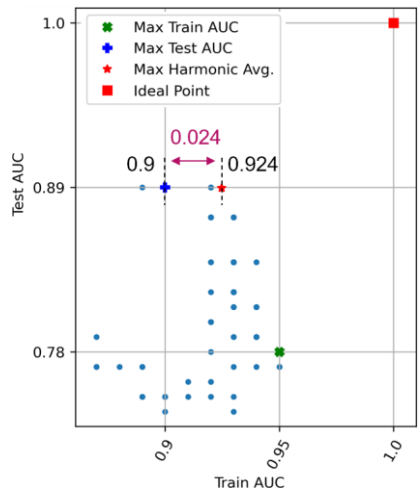

Fig. 4: Scatter of train vs test sets *AUC*, grid search of two hyperparameters: Threshold and number of principal components followed by 10 different classifiers.

Figs. 3 and 4 depict the scatter results of *accuracies* and *AUCs* for the train versus test sets. Table V shows more information about thresholds, PCs, and classifiers for gaining maximum values for different metrics.

*C. Dimensionality*

Excluding our encoder, Table III shows the dimensionality output of different category encoders which varies from 3 to 84. We sort the results of each one using different classifiers from Table V. The results based on maximum test *accuracy* shows that the available encoders with higher dimensionality output have more chances for higher *accuracy* results. Our method with only one output dimensionality using an SVM classifier defeats all of the other encoders. Prior researchers usually consider the number of PCs that capture 95% or 97% of the variance in the train set for dimensionality reduction problems which means they consider the variance dependency on of the PCs they define. Our results reveal that the number of PCs directly affects the output performance independently of the variance they capture and they should be considered a hyperparameter.

IV. CONCLUSION

This paper proposed a new method for converting categorical to numerical features, which can be adapted by choosing the correct threshold and number of Principal Components for different classifiers. Furthermore, it produced low dimensional outputs from high cardinality categorical variables. We used *accuracy* and *AUC* metrics to compare performances between our method and 17 available encoders. Additionally, we defined new metrics to estimate the trade-off between train and test set performances. Our results overcame the best encoder available for the *accuracy* test and our method achieved the same result for the *AUC* test with two floating points approximations. Data preparation and feature engineering are critical steps in every machine learning algorithm. Our encoder can contribute to achieving better performances. Our method involves data compression while translating categorical information, which could be useful in hybrid telecommunication networks such as 5G. Due to the power and resource constraints of IoT devices, our high-performance method may be an attractive solution for particular implementations. We can conclude that the new metrics provide a better trade-off between train and test performances with these results.

ACKNOWLEDGMENT

This research received funding from the European Union's Horizon 2020 research and innovation programme under the Marie Sklodowska-Curie Project Number 813391"